\documentclass[conference]{IEEEtran}
\IEEEoverridecommandlockouts
\usepackage{cite}
\usepackage{hyperref}
\usepackage{subcaption}
\usepackage{graphicx}
\usepackage{lipsum}
\usepackage{amsmath,amssymb,amsfonts}
\usepackage{graphicx}
\usepackage{textcomp}
\usepackage{xcolor}

\usepackage{multirow}
\usepackage{algorithm,algpseudocode}
\newcommand{\orcid}[1]{\href{https://orcid.org/#1}{\textcolor[HTML]{A6CE39}{\aiOrcid}}}
\usepackage{orcidlink}

\begin{document}

\title{TwinLiteNet: An Efficient and Lightweight Model for Driveable Area and Lane Segmentation in Self-Driving Cars\\
}
\author{
\IEEEauthorblockN{Quang-Huy Che\textsuperscript{\orcidlink{0009-0007-7477-4702}1, 2},Dinh-Phuc Nguyen\textsuperscript{1, 2}, Minh-Quan Pham\textsuperscript{1, 2},Duc-Khai Lam\textsuperscript{1, 2}}

\IEEEauthorblockA{\textsuperscript{1}University of Information Technology, Ho Chi Minh City, Vietnam}
\IEEEauthorblockA{\textsuperscript{2}Vietnam National University, Ho Chi Minh City, Vietnam}

Email:huycq@uit.edu.vn, 20521766@gm.uit.edu.vn, \{quanpm, khaild\}@uit.edu.vn
}
\maketitle

\begin{abstract}
Semantic segmentation is a common task in autonomous driving to understand the surrounding environment. Driveable Area Segmentation and Lane Detection are particularly important for safe and efficient navigation on the road. However, original semantic segmentation models are computationally expensive and require high-end hardware, which is not feasible for embedded systems in autonomous vehicles. This paper proposes a lightweight model for the driveable area and lane line segmentation. TwinLiteNet is designed cheaply but achieves accurate and efficient segmentation results. We evaluate TwinLiteNet on the BDD100K dataset and compare it with modern models. Experimental results show that our TwinLiteNet performs similarly to existing approaches, requiring significantly fewer computational resources. Specifically, TwinLiteNet achieves a mIoU score of 91.3\% for the Drivable Area task and 31.08\% IoU for the Lane Detection task with only 0.4 million parameters and achieves 415 FPS on GPU RTX A5000. Furthermore, TwinLiteNet can run in real-time on embedded devices with limited computing power, especially since it achieves 60FPS on Jetson Xavier NX, making it an ideal solution for self-driving vehicles. Code is available at \url{https://github.com/chequanghuy/TwinLiteNet}.

\end{abstract}


\section{Introduction} \label{intro}

Self-driving cars have emerged as a promising field in recent years, potentially revolutionizing transportation and road development. One crucial component of autonomous driving is perceiving the environment accurately and efficiently. Deep learning has been applied to real-world driving control tasks \cite{golf}\cite{tesla}. Semantic segmentation is a fundamental task in this problem, involving labeling each pixel in an image with a corresponding semantic class, such as road, vehicle, pedestrian, etc. This information can help autonomous cars navigate safely and avoid obstacles. Specifically, accurately detecting drivable areas and lane markings provides critical information for the system to make steering and lane-changing decisions. However, models like UNet \cite{unet}, SegNet \cite{segnet}, ERNet \cite{enet} for semantic segmentation; LaneNet \cite{lanenet} and SCNN \cite{scnn}, ENet-SAD \cite{enetsad} for lane line detection; YOLOP \cite{yolop}, YOLOPv2 \cite{yolopv2}, HybridNets \cite{hybridnets}, DLT-Net \cite{dltnet}, Multinet \cite{multinet} for multi-task problems often come with high computational costs and require high-end hardware, which is not suitable for embedded systems used in low-computational power autonomous vehicles.

This paper introduces a lightweight architecture that can be easily deployed on autonomous vehicle systems. The main contributions of our report are (1) A computationally efficient architecture for drivable area segmentation and lane detection. (2) Our proposed architecture, based on ESPNet \cite{espnet}, is an expandable convolutional segment network with Dilated Convolutions combined with Dual Attention Network \cite{cfam}, but instead of using a single decoder block, our TwinLiteNet utilizes two decoder blocks similar to YOLOP \cite{yolop}, YOLOPv2 \cite{yolopv2} for each task (3) Our experimental results demonstrate that TwinLiteNet achieves comparable performance with fewer parameters for various image segmentation tasks.


The remainder of the paper is represented in the sequence listed below: We evaluate relevant models in Section \ref{related} to grasp the benefits and drawbacks of modern models, such as Driveable Area Segmentation, Lane Detection, and Multi-task approaches so that we can benefit from their strengths and minimize the weaknesses of our model. The proposed TwinLiteNet, based on the preceding analysis, presents an architecture with methods to boost model performance in Section \ref{Propose}. The model we suggest strives for high processing speed and straightforward implementation on embedded computers. In Section \ref{result}, we evaluate TwinLiteNet and contrast it with other models performing the same task to determine how well it does. In the final Section, we provide some conclusions and future development directions.

\section{Related work} \label{related}
\subsection{Drivable Area Segmentation}

Semantic segmentation has been extensively researched in computer vision, and many efficient models have been developed for semantic segmentation in self-driving or object segmentation tasks. Specifically, for segmenting drivable areas, recent works have proposed efficient models that can achieve accurate results with low computational costs. ENet \cite{enet} is a lightweight CNN model that can run on embedded devices with limited resources in real-time. In the research \cite{under}, the authors have found using the Hybrid Dilated Convolution module to extract better feature representations from the input images for the segmentation task. Zhao et al.\cite{pspnet} devised the PSPNet model, which utilizes the Pyramid Pooling Module (PPM) that applies global average pooling with multiple different bin scales to extract features. Alongside complex computational models, Mehta et al. \cite{espnet} have proposed ESPNet with low computational costs, using Dilated Convolutions to build an efficient spatial pyramid (ESP) module. In addition to developing and presenting new models, Dual Attention Modules \cite{cfam} is proposed to enhance feature fusion.

\subsection{Lane Detection}

Segmentation based on deep learning is a practical approach for many studies on lane detection. Dual branch features are combined to perform the segmentation of lane versions. Although time-consuming, convolutional slicing of SCNN \cite{scnn} allows information to propagate across pixels along rows and columns within a layer. On the other hand, Enet-SAD \cite{enetsad} utilizes a self-attention-guided filter method to assist low-level feature maps in learning from high-level feature maps. In addition to segmentation, in recent years road markings \cite{pinet}\cite{ultra2} have also brought much attention from the community.
 \subsection{Multi-task Approaches}
 Multi-task learning is a popular approach to simultaneously address multiple tasks, allowing the model to learn shared representations and leverage commonalities across different tasks. A standard method for multi-task learning is to use a shared backbone network and separate heads for each task. For example, Mask R-CNN \cite{mask} is a model combining object detection and instance segmentation using a shared backbone network and different charges for each task. Similarly, LaneNet \cite{lanenet} is a model combining lane detection and lane segmentation using a shared backbone network and separate heads for each task. MultiNet \cite{multinet} accomplishes three tasks simultaneously: image classification, object recognition, and region segmentation for autonomous driving. To facilitate the exchange of information between tasks, DLT-Net \cite{dltnet} inherits the encoder-decoder structure and establishes cross-modal attention maps between task-specific decoder modules. \cite{geo} proposes inter-connected sub-structures between lane region segmentation and lane boundary detection. Additionally, it offers a unique minimization function to constrain lane boundaries to overlap with the lane region geometrically. Driveable Area Segmentation and Lane Detection are all tasks accomplished by Hybridnets \cite{hybridnets} using a shared encoder and three separate decoders. Recently, YOLOPv2 \cite{yolopv2} have proposed to use Bag-of-Freebies methods to achieve high accuracy and fast processing speed.

\section{Proposed method} \label{Propose}

In this section, we develop the lightweight model in detail. First, we first propose a model with an architecture of one encoder and two decoders for two separate tasks.; our TwinLiteNet consists of one input and two outputs so that the model learns the representation of two different tasks. We then recommend a Dual Attention Module to improve model performance. In addition, this section also proposes some loss functions used to train the model. We also present the training and inference mechanisms we use. The section below demonstrates our proposed method in detail.




\subsection{Model architectures}
\begin{figure*}[]
\centering
\includegraphics[width=0.98\linewidth]{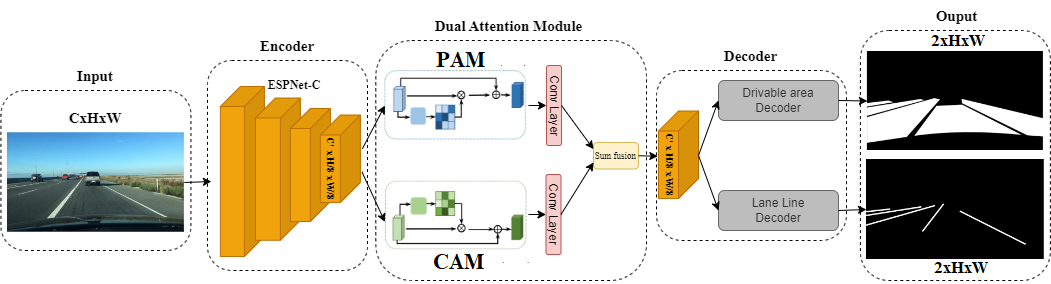}
\caption{The detailed TwinLiteNet architecture}    
\label{arch}

\end{figure*}

This paper presents a cost-effective architecture designed for task segmentation called TwinLiteNet, as illustrated in Figure \ref{arch}. Our approach utilizes ESPNNet-C as an information encoding block, enabling efficient feature map generation. We incorporate Dual Attention Modules into the network to capture global dependencies in both spatial and channel dimensions. These modules enhance the network's ability to perceive contextual information. The resulting feature map is then fed through two encoder blocks dedicated to performing two specific tasks: Driveable Area Segmentation and Lane Detection. By employing this architecture, we aim to achieve accurate and efficient segmentation results for these tasks at a reduced cost.

\begin{figure}[htp]
\centering
\includegraphics[width=0.8\linewidth]{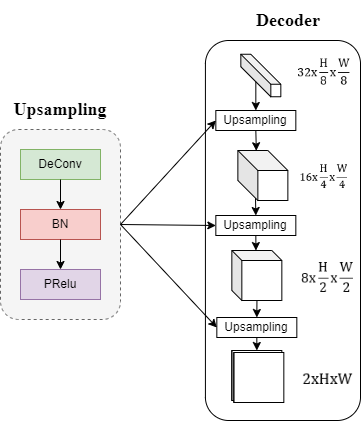}
\caption{Design of the proposed encoder block for the TwinLiteNet}    
\label{decode}

\end{figure}

Firstly, unlike approaches using backbones and high computational cost methods, we leverage the power of ESPNet with low computational cost but high accuracy. We use ESPNet-C as an encoder to extract features from the input image. In ESPNNet-C, in addition to sharing information through feature maps between ESP modules, it also consolidates input information at different dimensions between blocks in the architecture. After obtaining the feature map $A \in \mathbb R^{32 \times \frac{H}{8} \times \frac{W}{8}}$ from ESPNet-C, we pass the extracted features through Dual Attention Modules, we pass the extracted features through Dual Attention Modules  \cite{cfam}. The Dual Attention Modules consist of the Position Attention Module (PAM) and the Channel Attention Module (CAM). The PAM module is designed to incorporate a wider range of contextual information into local features, enhancing their representation capabilities. On the other hand, the CAM module leverages the interdependence among channel maps to highlight the interdependencies of feature maps and strengthen the representation of specific semantics. We transform the outputs of two attention modules by a Convolutional layer and employ an element-wise sum operation to achieve feature fusion $B \in \mathbb R^{32 \times \frac{H}{8} \times \frac{W}{8}}$. Our paper proposes a multi-output design for the Driveable Area and Lane segmentation tasks. Instead of using a single output for all object types requiring segmentation, we employ two decoder blocks to process the feature map and obtain the final results for each task. We recommend this multi-output design for the following reasons:\begin{itemize}
    \item Independent Performance Optimization: With two dedicated output blocks, we can optimize the segmentation performance independently for each class. This approach allows us to fine-tune and improve the segmentation results for Driveable Areas and Lanes separately without being influenced by the other class.
    \item Enhanced Accuracy: Using two output blocks for separate layers also improves segmentation accuracy. By focusing on each layer independently, our model can better learn and adjust the features specific to Driveable Areas and Lanes, leading to more accurate segmentation results for each class.
\end{itemize}
By adopting a multi-output design with two separate outputs for the Driveable Areas and Lanes segmentation tasks, we achieve independent performance optimization and enhanced segmentation accuracy for each class. Our decoder blocks are designed to be simple but ensure efficiency, relying on ConvTranspose layers followed by batch normalization and the pRelu \cite{prelu} activation function, as shown in Figure \ref{decode}. After decoding, TwinLiteNet returns two segmented images for the Driveable Area and Lane Detection tasks. Our TwinLiteNet optimizes the segmentation performance with high accuracy for both driveable area segmentation and lane detection tasks. By utilizing ESPNet-C and the feature analysis blocks Dual Attention Network, we enhance the feature extraction capability of the model. Besides, the simple decoder blocks help reduce computational costs and improve the model's efficiency. The output of the backbone and dual block has a small size of $32\times \frac{H}{8}\times \frac{W}{8}$, so the calculation of the subsequent blocks is optimized, leading to low inference time, and at the same time, Section \ref{result} is also proven with the proposed size. The model still carries enough information for learning and prediction.

\subsection{Loss function}

We utilize two loss functions for the proposed segmentation model: Focal Loss \cite{focal} and Tversky Loss \cite{tversky}. Focal loss aims to reduce the classification error among pixels while addressing the impact of easily predictable samples and heavily penalizing hard-to-predict samples, as shown in equation \ref{focal}. On the other hand, Tversky Loss draws inspiration from Dice Loss \cite{dice} and addresses the class imbalance issue in segmentation tasks. However, unlike Dice Loss, Tversky Loss introduces $\alpha$ and $\beta$ parameters to adjust the importance of false positives and false negatives during the computation, as described in the Tversky equation \ref{tversky}.

\begin{equation}
\label{focal}
    Loss_{focal}=-\frac{1}{N}\sum_{c=0}^{C-1}\sum_{i=1}^{N}p_i(c)(1-\hat{p}_i(c))^\gamma log(\hat{p}_i(c))
\end{equation}
\begin{itemize}[where:]
    \item $N$: A number of pixels in the input image
    \item $C$: A number of classes, in this case, one class is Drivable
Area or Lane, and the remaining class is the background.
    \item $\hat{p}_i(c)$: is to determine the prediction for pixel $i$ of class $c$
    \item $p_i(c)$: is the ground truth value for pixel $i$ of class $c$.
    \item $\gamma$: balance correction factor.
\end{itemize}
\begin{equation}
\label{tversky}
    Loss_{tversky}=\sum_{c=0}^{C}(1-\frac{TP(c)}{TP(c)-\alpha FN(c)-\beta FP(c)})
\end{equation}
\begin{itemize}[where:]
    \item $TP$: True Positives
    \item $FN$: False Negatives 
    \item $FP$: False Positives
    \item $C$: Number of classes, in this case, one class is Drivable Area or Lane, and the remaining class is the background.
    \item $\alpha$, $\beta$: Control the magnitude of penalties for FPs and FNs.
\end{itemize}

The aggregated loss function for each head take the form of the following:

\begin{equation}
\label{losstotal}
\begin{split}
    Loss_{total}=Loss_{focal}+Loss_{tversky}
\end{split}
\end{equation}
\subsection{Training mechanism and inference mechanism}
We trained our TwinLiteNet with input images of size 640x360. We used the Adam \cite{adam} optimizer with a decreasing learning rate over epochs. TwinLiteNet was trained in 100 epochs with a batch size of 32. During the inference process, we applied the Re-parameterization technique to merge the Convolution and Batch Normalization\cite{bn} layers into a single layer, which accelerated the inference speed. This merging process only occurs during inference, while during model training, they still operate as separate layers: Convolution and Batch Normalization. 


\section{Experimental results} \label{result}
The BDD100K \cite{bdd} dataset was used for training and validating TwinLiteNet. With 100,000 frames and annotations for 10 tasks, it is a large dataset for autonomous driving. Due to its diversity in geography, environment, and weather conditions, the algorithm trained on the BDD100K dataset is robust enough to generalize to new settings. The BDD100K dataset is divided into three parts: a training set with 70,000 images, a validation set with 10,000 images, and a test set with 20,000 images. Since no labels are available for the 20,000 images in the test set, we chose to evaluate on a separate validation set of 10,000 images. We resized the images in the BDD100k dataset from 1280$\times$720$\times$3 to 640$\times$360$\times$3. The preprocessing steps described in this study were conducted according to \cite{enetsad}. We used TorchScript to convert our model to a statically typed representation, which resulted in a speedup of inference without sacrificing accuracy.

All experiments used the PyTorch framework on an NVIDIA GeForce RTX A5000 GPU with 32GB of RAM and an Intel(R) Core(TM) i9-10900X processor.

\subsection{Cost Computation Performance}


Table \ref{tab:table0} compares the proposed model with other multi-task networks\footnote{These models also include task object detection.}. Our TwinLiteNet has only 0.4M parameters, much lower than previous models. Besides, our TwinLiteNet achieves 415FPS, while other models only achieve inference speed below 100FPS on the same test device.

\begin{table}[]
\begin{center}
    \centering
\caption{Computation Performance.}
\label{tab:table0}
\begin{tabular}{ccc}
\hline
\textbf{Network}    & \textbf{Params} & \textbf{Speed (FPS)} \\ \hline
YOLOP \cite{yolop}     & 7.9M   & 93                                                           \\
YOLOPv2 \cite{yolopv2}   & 38.9M  & 95                                                           \\
HybridNets \cite{hybridnets} & 13.8M  & 25                                                         \\ \hline
TwinLiteNet (our)        & \textbf{0.4M}   & \textbf{415}                                                        \\ \hline                                       
\end{tabular}
\end{center}
\end{table}

\subsection{Drivable Area Segmentation Result}

In this paper, both the ``drivable area'' and ``alternative area'' in the BDD100K dataset were converted to ``drivable area''. The mean Intersection over Union (mIoU) metric was used to evaluate the segmentation performance of our TwinLiteNet. The results are shown in Table \ref{tab:table1}. Although the accuracy of TwinLiteNet is higher than some previously proposed models, our TwinLiteNet is slightly lower than the current State-of-the-Art (SOTA) YOLOPv2 (-1.88\%) and lower than YOLOP (-0.18\%). It is easy to understand because TwinLiteNet is developed to optimize inference time rather than accuracy. Some segmentation results of the model are shown in Figure \ref{DA}. The results showcase the impressive performance of TwinLiteNet in accomplishing the task under diverse lighting conditions, including both daytime and nighttime scenarios. Notably, the network demonstrates high accuracy in segmenting the Driveable Area while effectively avoiding confusion with the opposite lane. 

\subsection{Lane Detection Result}

\begin{figure*}[h!]
  \centering
    \begin{subfigure}{\columnwidth}
        \includegraphics[width=0.95\textwidth]{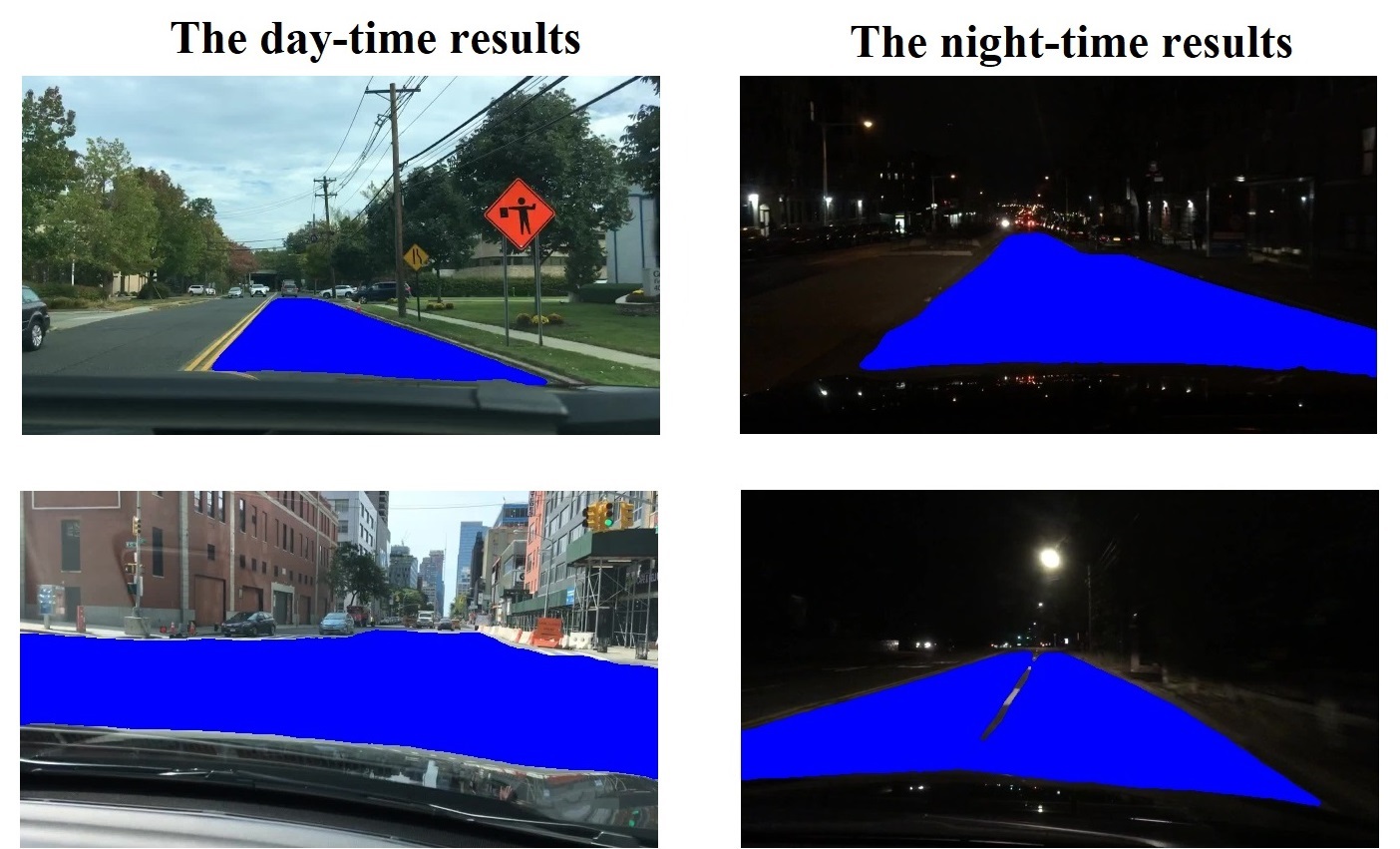}
        \caption{Drivable area segmentation} 
        \label{DA}
    \end{subfigure} 
    \hfill 
    \begin{subfigure}{\columnwidth}
        \includegraphics[width=1.\textwidth]{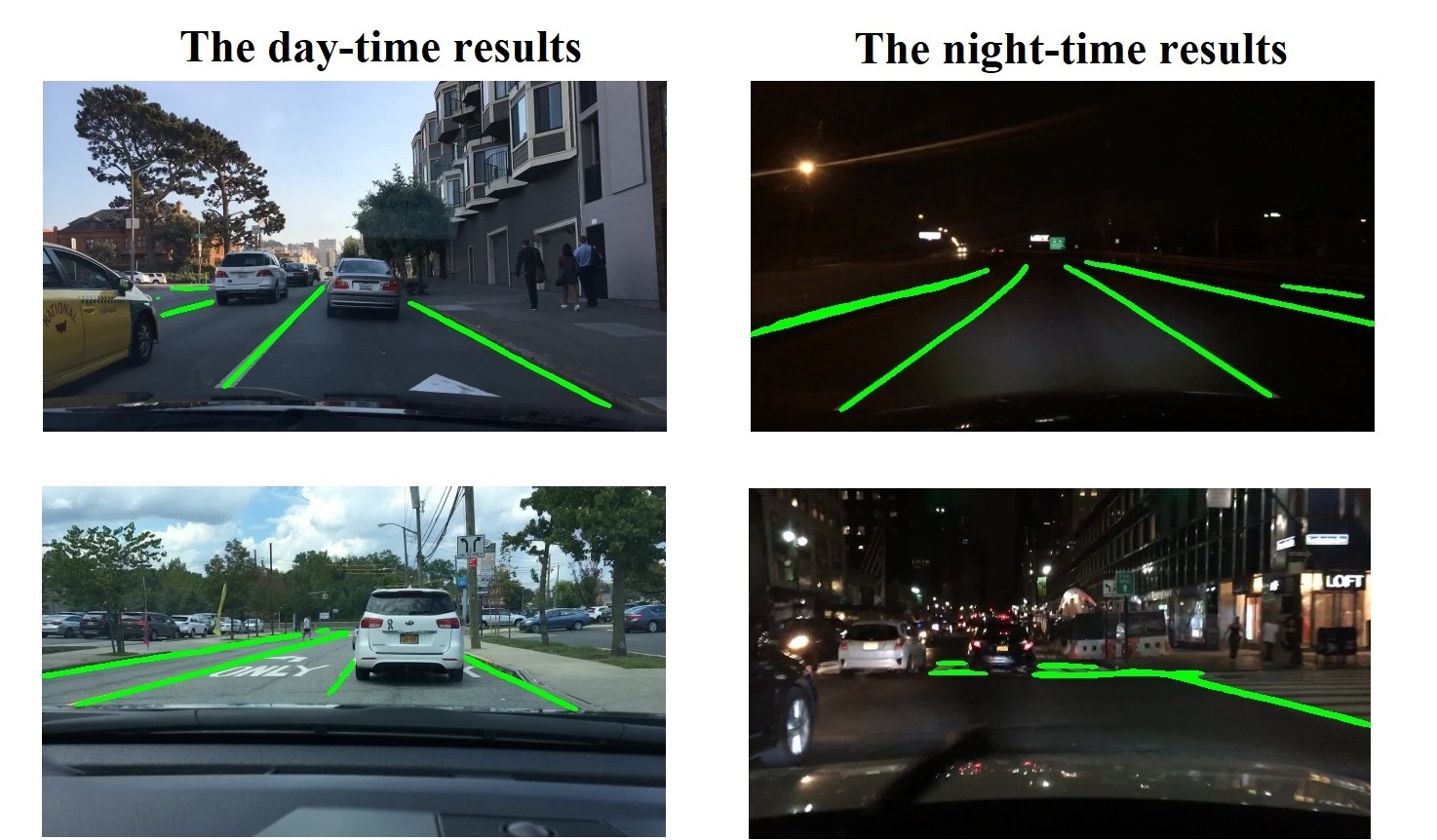}
        \caption{Lane detection} 
        \label{LL}
    \end{subfigure} 
    \caption{Visualization of the results}
\end{figure*}

The lane lines in the BDD100K dataset are labeled using lines with low pixel thickness. To compare expediently with previous research, following \cite{enetsad}, we adopted the methodology of merging two-lane line annotations into a single central line during the annotation process. In the training set, we applied dilation to expand the annotations by 8 pixels so that the model could learn better while keeping the validation set the same to ensure consistency with previous and future research. We use IoU (Intersection over Union) of lane lines as evaluation metrics. As listed in Table \ref{tab:table2}, TwinLiteNet is still lower than the current SOTA HybridNets in terms of IoU (-0.52\%). Some segmentation results of the model are shown in Figure \ref{LL}. Our experimental findings reveal that our model exhibits strong predictive capabilities for multi-lane roads, performing admirably in daytime and nighttime scenarios. This observation underscores the model's ability to accurately anticipate and predict lane configurations, regardless of the lighting conditions.
\begin{table}[]
\begin{center}
    \centering
\caption{Drivable Area Segmentation Results.}
\label{tab:table1}
\begin{tabular}{cc}
\hline
\textbf{Network}    & \textbf{mIoU (\%)}       \\ \hline
MultiNet \cite{multinet}  & 71.6          \\
DLT-Net  \cite{dltnet}  & 71.3          \\
PSPNet  \cite{pspnet}   & 89.6          \\
HybridNets \cite{hybridnets} & 90.5          \\
YOLOP   \cite{yolop}   & 91.5          \\
YOLOPv2  \cite{yolopv2}  & \textbf{93.2} \\ \hline
TwinLiteNet (our)       & 91.3 (-1.9) \\ \hline
\end{tabular}
\end{center}
\end{table}


\begin{table}[]
\begin{center}
    \centering
\caption{Lane Detection Results.}
\label{tab:table2}
\begin{tabular}{cc}
\hline
\textbf{\textbf{Network}}    & \textbf{\textbf{IoU (\%)}} \\ \hline
ENet \cite{enet}             & 14.64                      \\
SCNN  \cite{scnn}            & 15.84                      \\
R-101-SAD \cite{enetsad}     & 15.96                      \\
ENet-SAD \cite{enetsad}      & 16.02                      \\
YOLOP  \cite{yolop}          & 26.20                      \\
YOLOPv2 \cite{yolopv2}       & 27.25                      \\
HybridNets \cite{hybridnets} & \textbf{31.6}              \\ \hline
TwinLiteNet (our)            & 31.08 (-0.52)              \\ \hline
\end{tabular}
\end{center}
\end{table}



\subsection{Ablation studies}

In this section, we investigate the ablation study. The ablation options we proposed and their corresponding results are presented in Table \ref{tab:table4}. We evaluate the results from a simple baseline that only includes the backbone and one output for the whole Drivable Area as Lane Detection. Then we gradually add Dual Attention Module, Multiple Head, and Re-parameterization methods. The results show that with the baseline model, we achieve the highest inference speed with 530 FPS but at the expense of much accuracy compared to the fully integrated model, which achieves 415 FPS inference speed.

\begin{table*}[]
\begin{center}
\caption{Ablation studies}
\label{tab:table4}
\begin{tabular}{ccc|cccc}
\hline
\multirow{2}{*}{\textbf{Dual Attention Module}} & \multirow{2}{*}{\textbf{Multiple Head}} & \multirow{2}{*}{\textbf{Re-parameterization}} & \textbf{Drivable Area} & \textbf{Lane Detection}    & \multirow{2}{*}{\textbf{FPS}} & \multirow{2}{*}{\textbf{Parameter}} \\
                                                &                                         &                                      & \textbf{mIoU (\%)}     & \textbf{IoU (\%)} &                      &                            \\ \hline
                                                &                                         &                                      & 88.7                   & 24.7              & \textbf{530}         & \textbf{417020}            \\
\checkmark                                      &                                         &                                      & 89.3                   & 25.6              & 425                  & 436966                     \\
\checkmark                                      & \checkmark                              &                                      & \textbf{91.3}          & \textbf{31.08}    & 400                  & 439633                     \\
\checkmark                                      & \checkmark                              & \checkmark                           & \textbf{91.3}          & \textbf{31.08}    & 415                  & 439339                     \\ \hline
\end{tabular}
\end{center}

\end{table*}

\subsection{Edge Devices}

\begin{figure*}[h]
  \centering
    \begin{subfigure}[b]{0.84\columnwidth}
    \centering
        \includegraphics[width=0.85\textwidth]{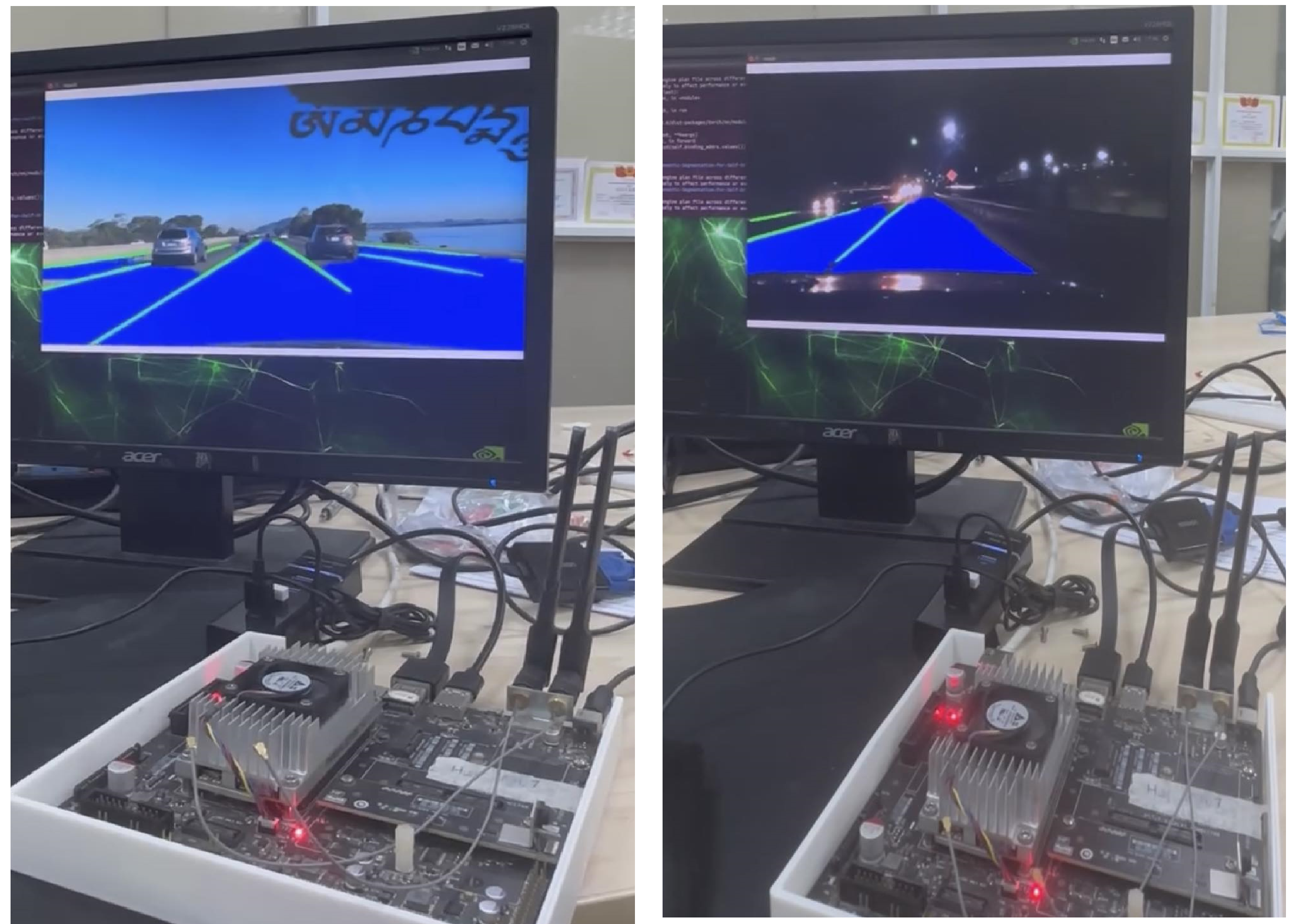}
        \caption{Jetson TX2} 
    \end{subfigure} 
    \begin{subfigure}[b]{0.85\columnwidth}
    \centering
        \includegraphics[width=0.85\textwidth]{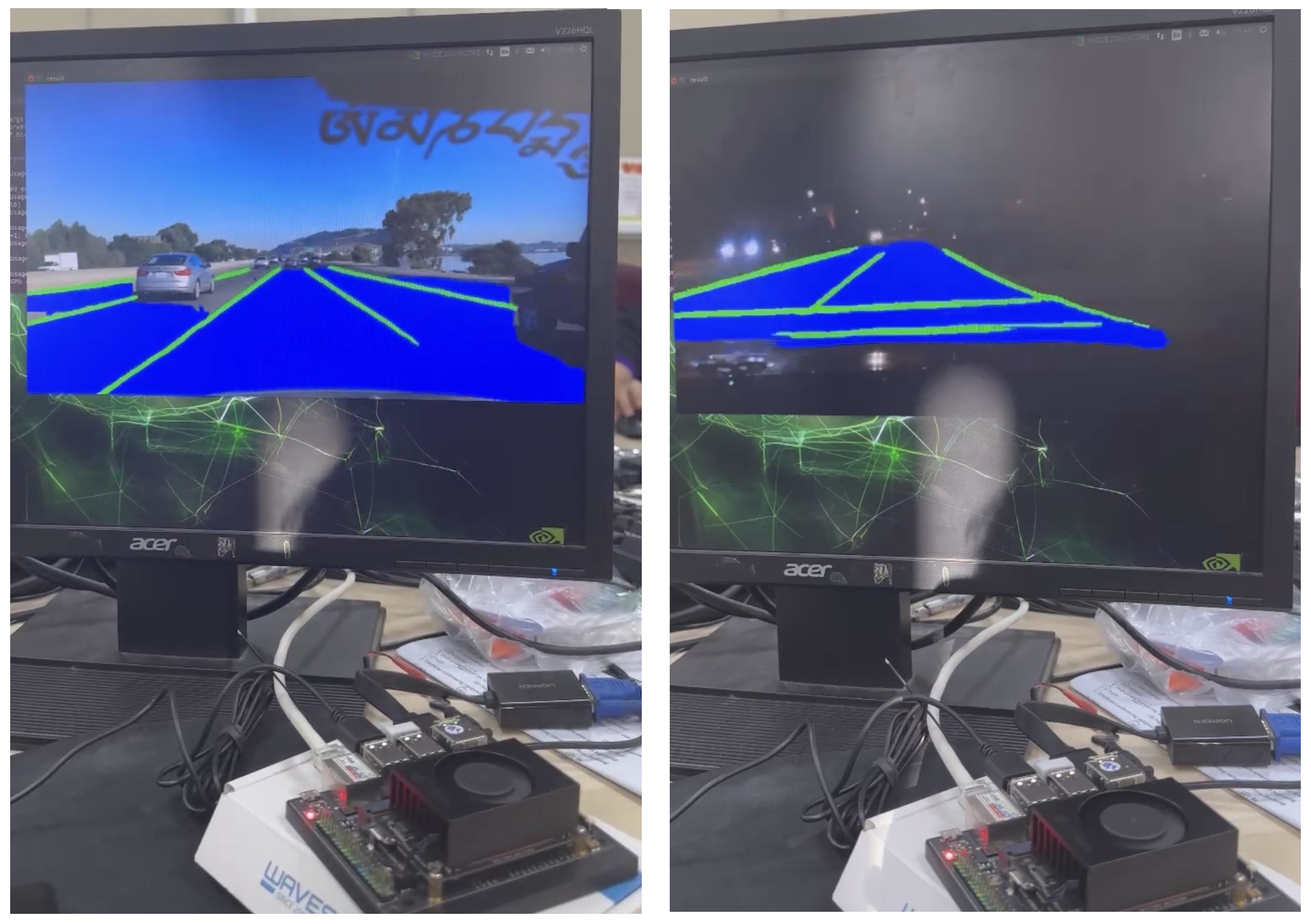}
        \caption{Jetson Xavier} 
        
    \end{subfigure} 
    \caption{Visualization of some results when implemented on the kit}
    \label{flipchar}
    \label{edge}
\end{figure*}

This section presents the inference speed results of deploying our TwinLiteNet model on embedded devices, specifically the NVIDIA Jetson TX2 and Jetson Xavier NX. We utilized the TensorRT SDK for performing model inference on these Jetson devices. The results demonstrate that our TwinLiteNet model achieves real-time computation on edge devices, with the Jetson Xavier NX achieving a frame rate of 60 FPS and the Jetson TX2 reaching 25 FPS. To evaluate the performance of our model, we recorded the prediction process on the edge devices\footnote{\url{https://youtube.com/shorts/TSklO0L4XJs}} using test videos from the BDD100K dataset with some visible results shown in Figure \ref{edge}. These results demonstrate that the TwinLiteNet model achieves accurate predictions in both daytime and nighttime images across various embedded devices.

\begin{figure}[h]
  \centering
    \begin{subfigure}[b]{0.48\columnwidth}
    \centering
    
        \includegraphics[width=\textwidth]{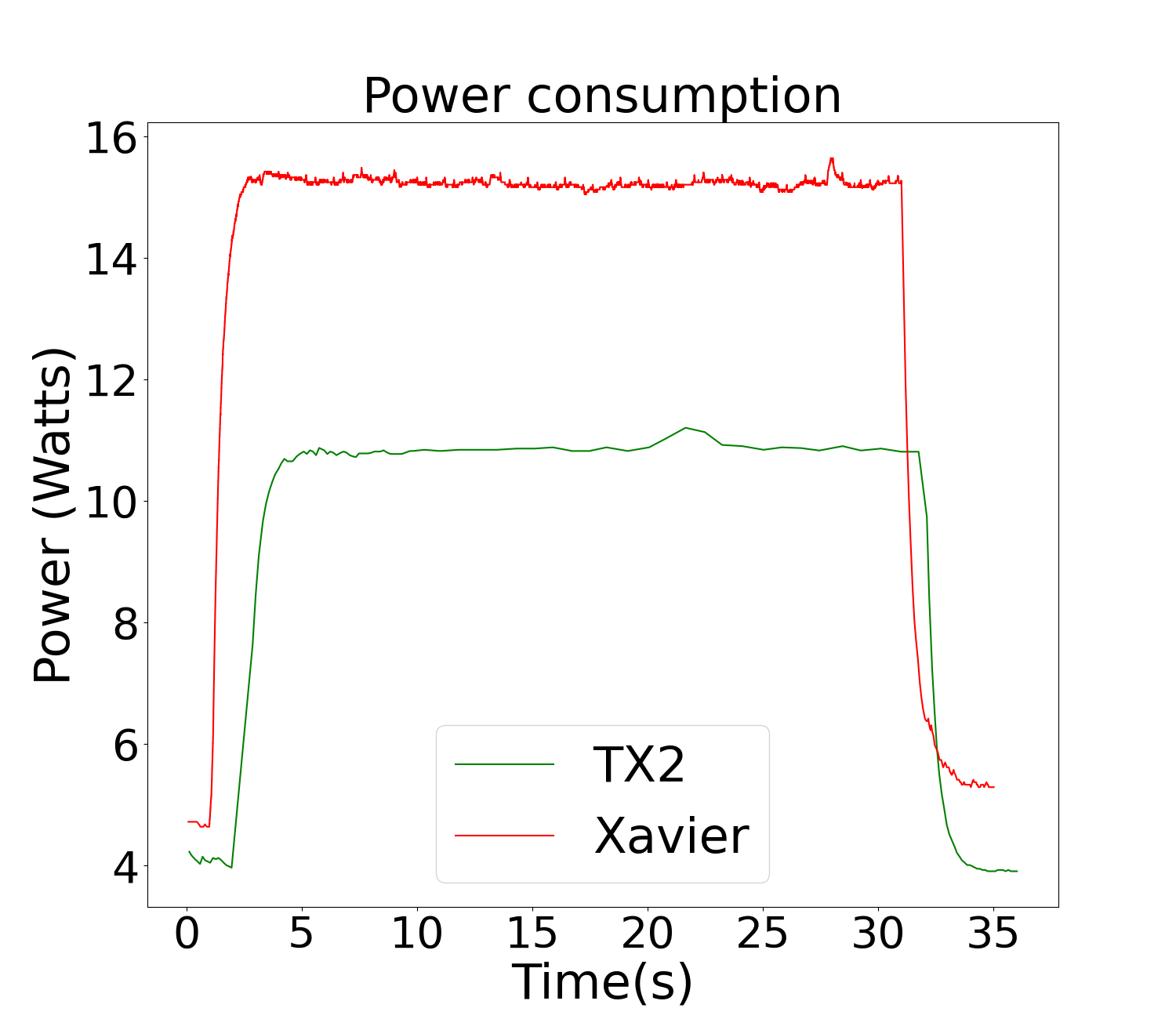}
        \caption{Power consumption} 
        \label{vol}
    \end{subfigure} 
    \begin{subfigure}[b]{0.48\columnwidth}
    \centering
        \includegraphics[width=\textwidth]{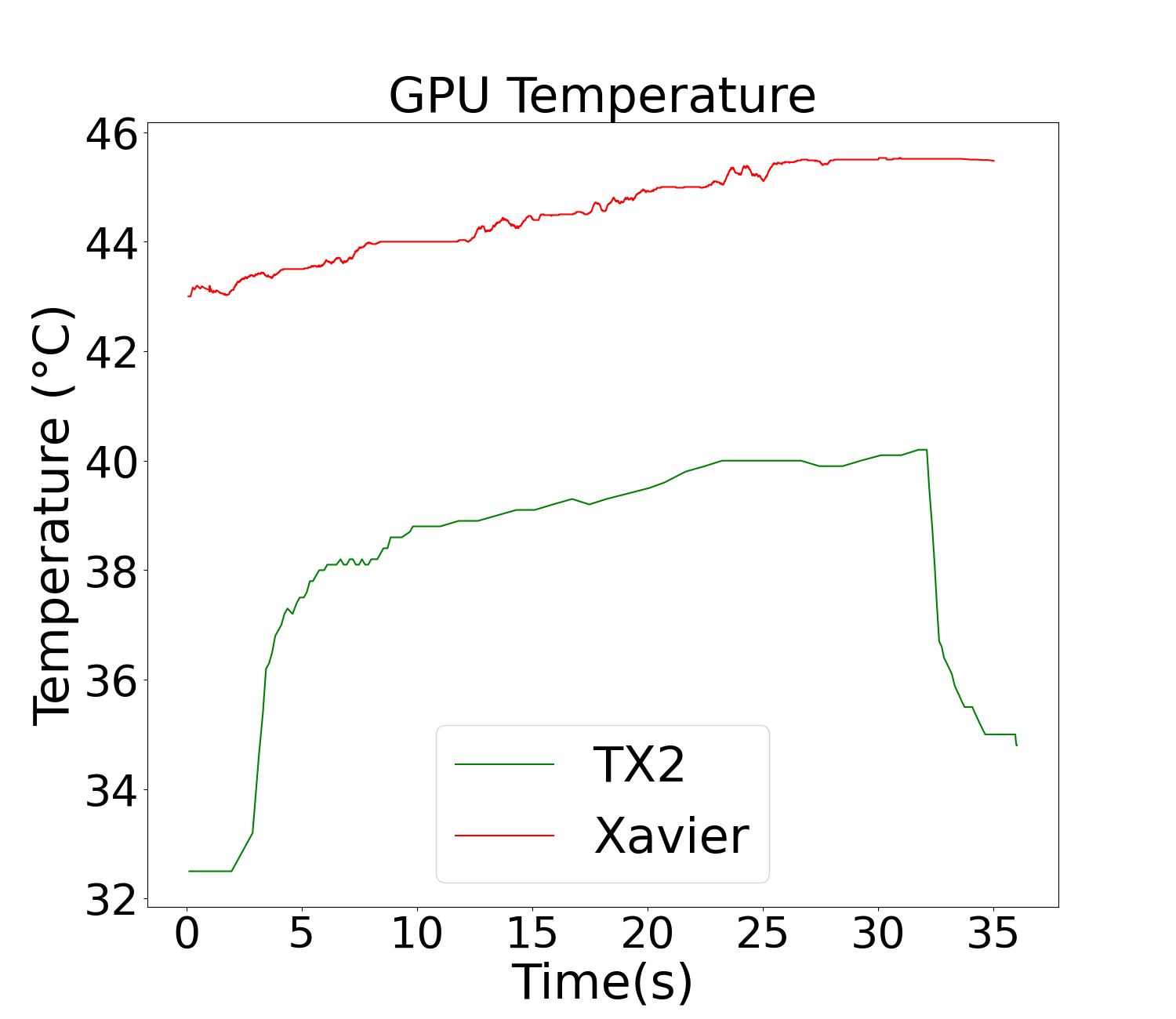}
        \caption{Temperature consumption} 
        \label{temp}
    \end{subfigure} 
    \caption{Consumption on TX2 vs. Xavier}
    
\end{figure}

Furthermore, we monitored the power consumption and operating temperature of the TwinLiteNet model on the embedded devices. Figure \ref{vol} illustrates the recorded power consumption, and Figure \ref{temp} depicts the operating temperature of the model during inference on the proposed edge devices. These measurements provide insights into the energy efficiency and thermal performance of our TwinLiteNet model in real-world deployment scenarios.

\section{Conclusion} \label{conclusion}

We introduce a lightweight and energy-efficient segmentation model for autonomous driving tasks, specifically Drivable Area and Lane Detection. Our TwinLiteNet aims to achieve high processing speed with a slight trade-off in accuracy. Evaluation of the BDD100K dataset demonstrates that our model delivers a good balance between accuracy and high computational speed on GPUs and even on edge devices. In the future, we intend to evaluate the performance of the TwinLiteNet model on various publicly available datasets and apply the model to real-world scenarios. This approach enables us to assess its effectiveness in different contexts and address practical challenges.

\section*{Acknowledgement}
This research is funded by Vietnam National University HoChiMinh City (VNU-HCM) under grant number DS2023-26-02.
{\small
\bibliographystyle{IEEEtran}
\bibliography{ref}
}

\end{document}